\documentclass[conference]{IEEEtran}
\IEEEoverridecommandlockouts
\usepackage{cite}
\usepackage{amsmath,amssymb,amsfonts}
\usepackage{algorithmic}
\usepackage{graphicx}
\usepackage{textcomp}
\usepackage{xcolor}

\def\BibTeX{{\rm B\kern-.05em{\sc i\kern-.025em b}\kern-.08em
    T\kern-.1667em\lower.7ex\hbox{E}\kern-.125emX}}
\begin{document}

\title{Learning robot motor skills with mixed reality\\
}

\author{\IEEEauthorblockN{Eric Rosen}
\IEEEauthorblockA{
\textit{Brown University}\\
eric\_rosen@brown.edu}
\and
\IEEEauthorblockN{ Sreehari Rammohan}
\IEEEauthorblockA{
\textit{Brown University}\\
sreehari\_rammohan@brown.edu}
\and
\IEEEauthorblockN{Devesh Jha}
\IEEEauthorblockA{
\textit{MERL}\\
jha@merl.com}

}

\maketitle

\begin{abstract}
Mixed Reality (MR) has recently shown great success as an intuitive interface for enabling end-users to teach robots. Related works have used MR interfaces to communicate robot intents and beliefs to a co-located human, as well as developed algorithms for taking multi-modal human input and learning complex motor behaviors. Even with these successes, enabling end-users to teach robots complex motor tasks still poses a challenge because end-user communication is highly task dependent and world knowledge is highly varied. We propose a learning framework where end-users teach robots a) motion demonstrations, b) task constraints, c) planning representations, and d) object information, all of which are integrated into a single motor skill learning framework based on Dynamic Movement Primitives (DMPs). We hypothesize that conveying this world knowledge will be intuitive with an MR interface, and that a sample-efficient motor skill learning framework which incorporates varied modalities of world knowledge will enable robots to effectively solve complex tasks.

\end{abstract}

\begin{IEEEkeywords}
robot learning, mixed reality, human-robot interaction
\end{IEEEkeywords}

\section{Introduction}
Mixed Reality (MR) has recently garnered interest as an intuitive interface for teaching robots. MR enables bi-directional human-robot interaction by superimposing visualizations on the shared environment to communicate robot knowledge and beliefs to the human, and providing a means for humans to communicate to the robot via multi-modal channels like speech, eye-gaze, and hand-gestures. There have been many recent studies that have demonstrated the efficacy and ease-of-use of MR interfaces for teaching robots compared to traditional methods like kinesthetic teaching and keyboard/monitor setups.

\begin{figure}[h]
\includegraphics[scale=0.22]{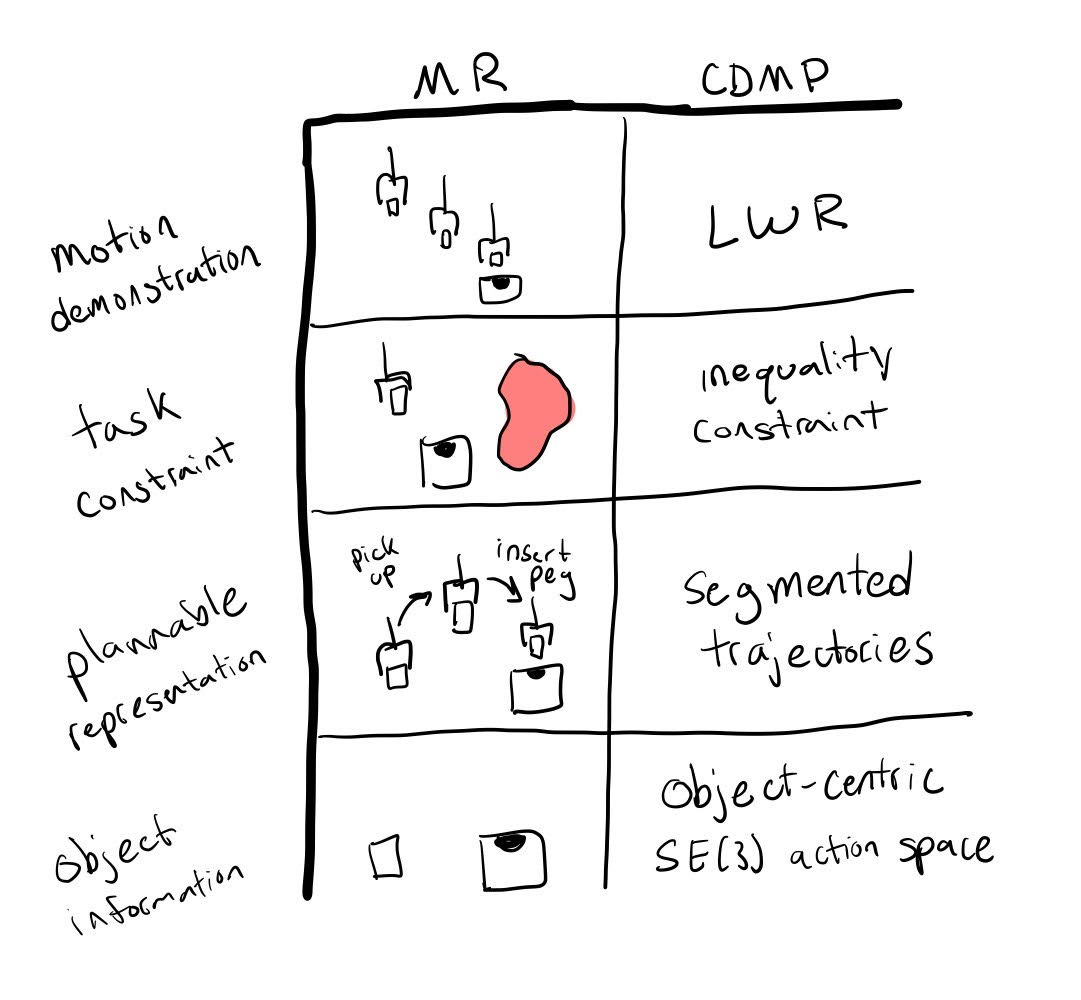}
\caption{A figure describing the types of world knowledge our proposed MR interface can enable users to provide, and how we propose a single motor skill learning framework (CDMPs) can leverage that world knowledge. More details on how each type of world knowledge is communicated with MR and integrated into the CDMP can be found in Section \ref{proposed}}
\label{cover}
\end{figure}

 Human communication is varied and task dependent, making robot teaching difficult. We identify 4 types of world knowledge a human may wish to impart: a) motion demonstrations, b) task constraints, c) planning representations, and d) object information. \textbf{We address the research problem of incorporating this world knowledge into a single motor skill learning framework with MR}. 

We propose that a suitable motor skill learning framework must be sample-efficient (to reduce the teaching burden from a human), provide guarantees on constraint satisfaction (for safety), and be adaptable to new goals (to account for novel task dependent constraints). We propose Constrained Dynamic Movement Primitives (CDMPs) as an appropriate family of motor skills for humans to teach using MR (Figure \ref{cover}). We will develop an MR interface for users to specify all the aforementioned types of world knowledge for a peg-in-hole insertion task, and demonstrate that all the information can effectively be incorporated into a CDMP learning framework to learn complex motor skills.

\section{Related work}
Related works within the VAM-HRI (Virtual, Augmented, and Mixed Reality for Human-Robot Interaction) have developed interfaces and learning algorithms for different types of world knowledge. Here we discuss both related works with MR interfaces for teaching different types of world knowledge, and also give background on Constrained DMPs.

\subsection{Mixed Reality for Robot Learning}
MR has been used in various contexts for robot learning. Here we discuss 4 different types of world knowledge humans have taught robots: motion demonstrations, task constraints, planning representations, and object information.

\textbf{Motion demonstrations}
Spatially-tracked hand controllers have shown great success as an intuitive interface to teleoperate the pose of a robot's end effector to perform manipulation tasks \cite{whitney2020comparing}. For robot learning with MR, users typically provide motion demonstrations by teleoperating the robot with spatially-tracked controller (often mapped to the end-effector with Cartesian control), and these teleoperation demonstrations are used as supervised data within a motor skill learning framework. \cite{zhang2018deep} demonstrated that commercially-available VR systems could be used to provide demonstrations of complex manipulation tasks to support effective imitation learning with deep neural networks. \cite{delpreto2020helping} used an apprenticeship model to efficiently use a human's time when teleoperating in VR to teach a grasping task.

\textbf{Task constraints}
Visualizing task constraints and enabling end-users to easily edit them with MR has recently shown success for positively augmenting the learning process. \cite{choi2022integrated} propose an MR system for safety-aware HRI where users' skeletons are registered and tracked to calculate safety distances between the human and robot, which are treated as task constraints that can be visualized to the user. Beyond visualizing constraints, \cite{luebbers2019augmented} proposed an AR system to teach robots motor skills using Concept-Constrained Learning from Demonstration (CC-LfD), allowing users to see and interact with visualizations of constraints superimposed on the workspace.
\cite{luebbers2021arc} expanded on CC-LfD by incorporating the ability to perform skill maintenance and skill alteration for different contexts. 

\textbf{Planning representations}
Tasks are often solved by decomposing the goal into subgoals, and related works have investigated creating plans comprised of high-level action sequences with MR. \cite{gadre2018teaching} used an MR system to segment demonstrations in MR with a key-frame system, and then fitted different motor policies with each of the sub-trajectories. \cite{fang2009robot} developed an AR system for users to specify trajectory planning information that leverages dynamics modeling to produce smooth trajectories between initial and end points for the robot.

\textbf{Object information}
MR provides an intuitive way for humans to communicate information about objects to a robot in a shared environment.
\cite{rosen2020building} proposed an MR interface for users to build a semantic map of an environment that included information such as object pose, semantic attributes and state, and action-related information like grasp strategies and approach vectors. \cite{puljiz2021hair} propose an AR system for intent recognition that enables users to directly specify 3D pose information about objects in the scene, which can be used to model and infer a human's intended goal object using Hidden Markov Models (HMM).

\begin{figure}
    \centering
    \includegraphics[width=0.5\linewidth
    ]{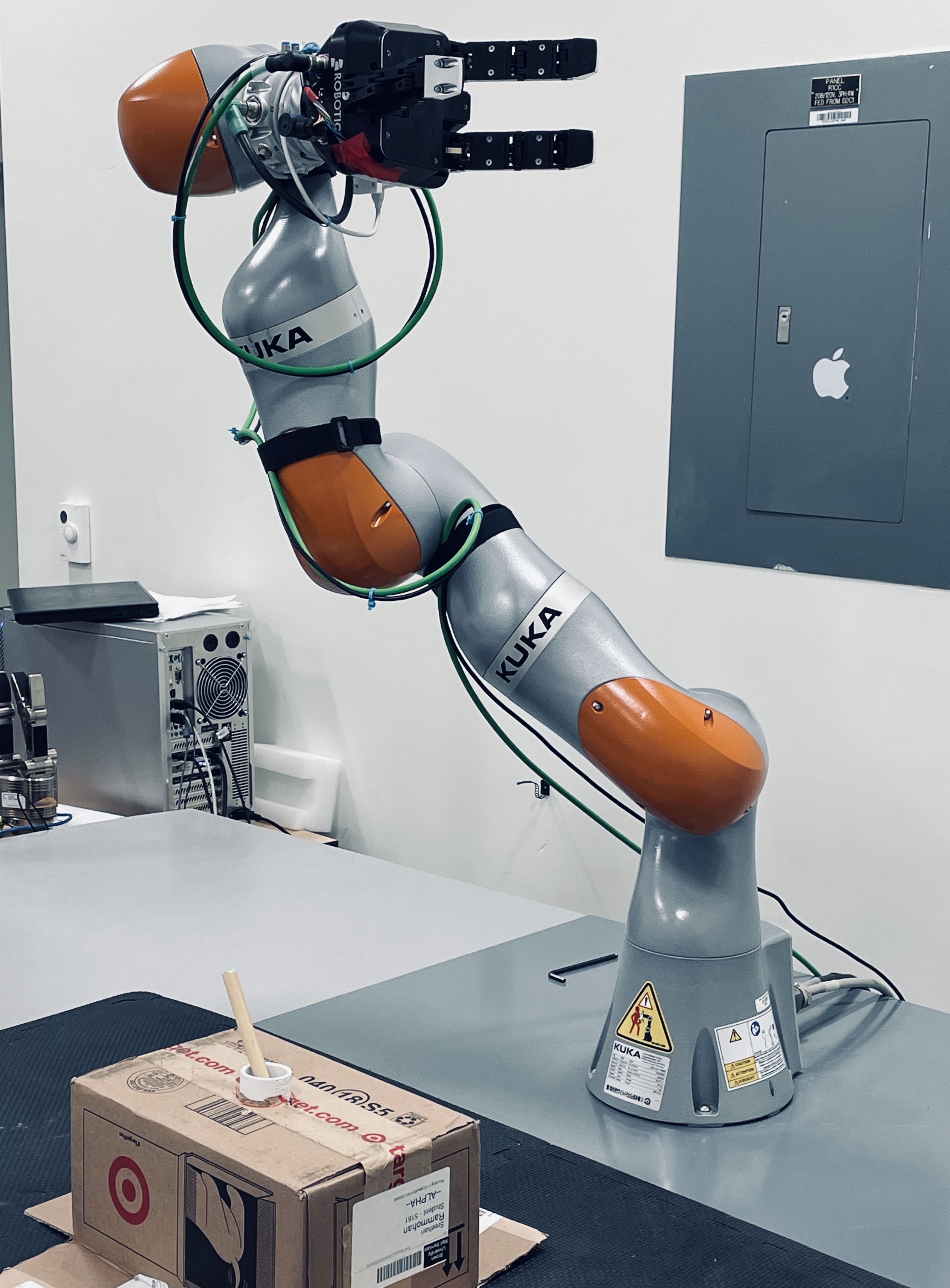}
    \caption{The Kuka Arm and Peg-In-Hole insertion task.}
    \label{fig:arm}
\end{figure}

\begin{figure}
    \centering
    \includegraphics[width=0.8\linewidth]{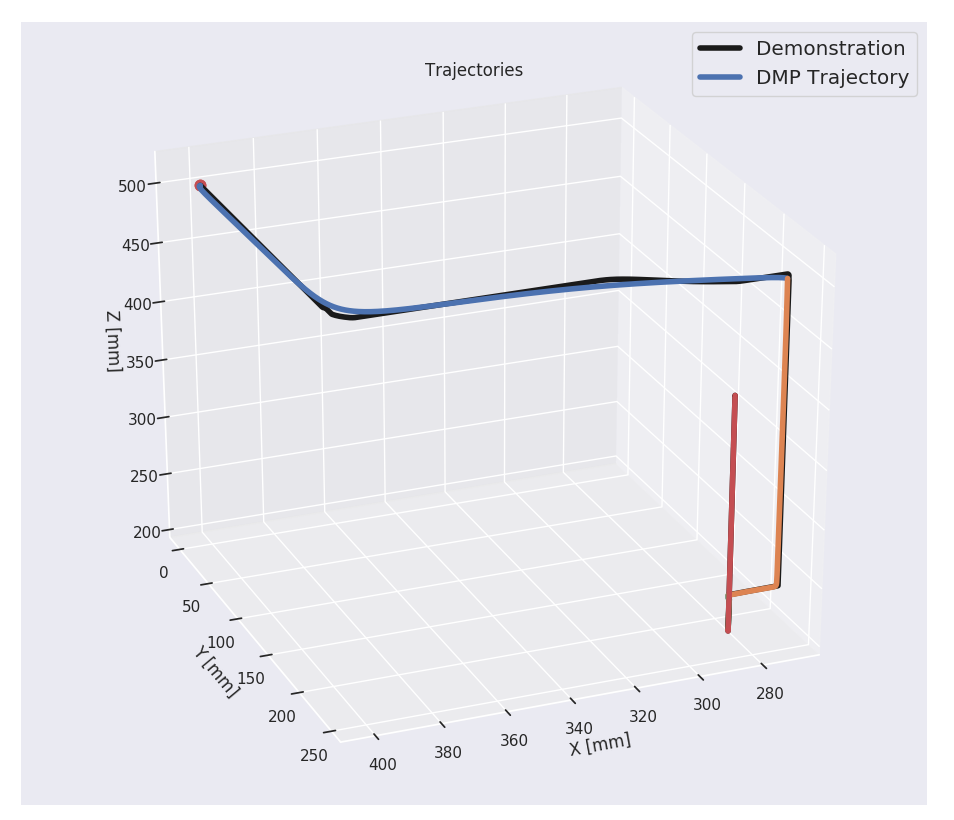}
    \caption{Insertion Trajectory for Peg in Hole Task}
    \label{fig:3dtraj}
\end{figure}

\subsection{Constrained Dynamic Movement Primitives (CDMPs)}
Constrained DMPs (CDMPs) were recently introduced in~\cite{shaw2022cdmp} as an extension of the original DMP formulation introduced in~\cite{schaal2006dynamic}. The idea in CDMP is to allow constraint satisfaction for the original DMP by introducing a perturbation in the learned forcing function. More concretely, the CDMP is represented by the following system of equations:
\begin{equation}
\label{eqn:tensor_dmp}
\begin{bmatrix}
   \dot s \\
   \dot z \\
   \dot y \\
\end{bmatrix}
=
\frac{1}{\tau}
\begin{bmatrix}
    -\alpha_s s\\
    \left(\alpha_z(\beta_z (g - y) - z) +\tilde{f}(s)\right)\\
    z \\ 
\end{bmatrix}
\end{equation}
where,
\begin{equation}
    \tilde{f}(s) = \frac{\sum_{i=1}^N (w_i - \zeta_i) \psi_i(s)}{\sum_{i=1}^N \psi_i(s)}
\end{equation}
where, $w_i$ are the weights of the forcing function learned using the expert demonstration and $\zeta_i$ are the weights learnt to satisfy the operational constraints by solving a constrained trajectory optimization problem. Note that the forcing function presented in~\cite{schaal2006dynamic} consists only of the weights $w_i$ that can be learned from the expert demonstration by solving a locally weighted regression~\cite{schaal2006dynamic}. In the formulation presented in~\cite{shaw2022cdmp}, constraints could be represented as barrier functions that guarantee safety or constraint satisfaction if a feasible solution could be found. These constraints could represent collision constraints or more complex constraints that could represent task-specific constraints. CDMP could be solved using existing nonlinear programming solvers~\cite{DBLP:journals/corr/abs-2106-03220}.
\section{Proposed Work}
\label{proposed}
\begin{figure}
    \centering
    \includegraphics[width=1\linewidth]{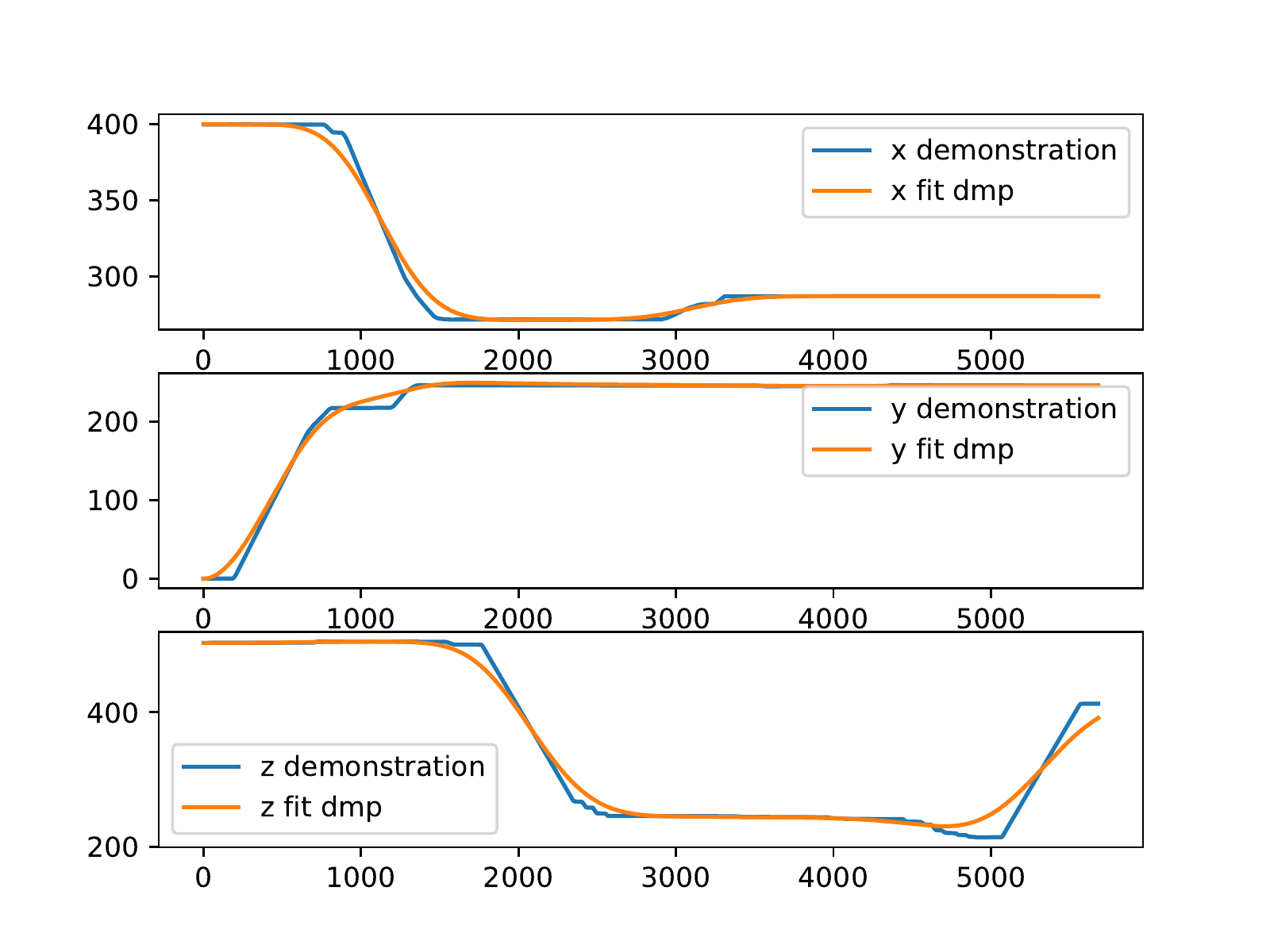}
    \caption{Fitted DMP for Peg-in-Hole insertion task.}
    \label{fig:fitted_DMP}
\end{figure}

Our proposed work is to build an MR interface that is able to take in the four types of aforementioned world knowledge, and apply it to the CMDP learning framework for enabling end-users to teach a robot how to perform a peg-in-hole insertion task (Figure \ref{fig:arm}). For supplying motion demonstrations, users will be able to teleoperate the robot's end effector by manipulating a spatially-tracked hand controller (Figure \ref{fig:3dtraj}). We will use these demonstrations to fit the forcing function of the CDMP using Locally Weighted Regression (LWR), which is both sample and computationally efficient (Figure \ref{fig:fitted_DMP}). For specifying task constraints, users will be able to specify and adjust geometric shapes (for now, spheres and rectangular prisms) over the workspace marking zones of space where the robot's end effector can not enter. The CDMP will incorporate these task constraints into the motor skill learning by automatically constructing an inequality using signed distance functions that respect the spatial constraints specified by the user, which provide guarantees on constraint satisfaction. For building plannable representations, users will be able to designate keypoints within a trajectory where skills begin and end. For each keypoint, we will segment the trajectory and train a separate CDMP on each subset trajectory. At inference time, these CDMPs are stitched together to chain skills and solve the given task. For specifying object information, users will be able to generate object models and directly overlay them on top of their real-world counterparts, and associate different motor skills with each of the objects. CMDPs will have their action-space defined in $SE(3)$, and will leverage the specified object information to determine what frame of references the action-space is in (either the global work-space or object-centric), which will in turn define what frame of reference reparameterized goals are with respect to.

Once we create our propsed MR interface and connect all the components to a CDMP implementation, we will conduct a user study to understand how easy it is for users to communicate these types of world knowledge, and how sufficient they are for enabling robots to learn complex motor works. We will have end-users try our robot teaching system with a peg-in-hole insertion task. This will give us a better understanding of whether or not there are additional types of world knowledge that should be incorporated into the learning framework. After, we can then compare the efficacy of our MR interface to traditional interface methods (such as kinesthetic teaching and desktop setups) for enabling end-users to teach robots a complex task such as peg-in-hole insertion.

\section{Conclusion}
We have proposed an MR interface and compatible motor learning framework for enabling end-users to teach robots manipulation skills. We hypothesize that MR can be an effective interface for expressing four types of world knowledge: motion demonstrations, task constraints, plannable representations, and object information. We suggest that CDMPs are an appropriate motor skill learning framework for learning these different types of world knowledge without requiring large numbers of interactions. Next steps will be to build the MR interface and connect it to the motor skill learning framework, and conduct a user study to compare MR against traditional robot interfaces for teaching peg-in-hole insertion.

\section{Acknowledgements} We express appreciation for the support from the Humans To Robots Laboratory and the Intelligent Robot Lab at Brown University for supplying our research with robotic and MR platforms.

\bibliographystyle{IEEEtran}
\bibliography{bibliography}

\end{document}